\documentclass{ieeeaccess}
\usepackage{cite}
\usepackage{amsmath,amssymb,amsfonts}
\usepackage{algorithm}
\usepackage{algpseudocode}
\usepackage{graphicx}
\usepackage{textcomp, nicefrac}
\usepackage{bm}
\usepackage{multirow, multicol, booktabs}
\usepackage{url}
\usepackage{hyperref}
\usepackage{balance}
\makeatletter

\AtBeginDocument{\DeclareMathVersion{bold}
\SetSymbolFont{operators}{bold}{T1}{times}{b}{n}
\SetSymbolFont{NewLetters}{bold}{T1}{times}{b}{it}
\SetMathAlphabet{\mathrm}{bold}{T1}{times}{b}{n}
\SetMathAlphabet{\mathit}{bold}{T1}{times}{b}{it}
\SetMathAlphabet{\mathbf}{bold}{T1}{times}{b}{n}
\SetMathAlphabet{\mathtt}{bold}{OT1}{pcr}{b}{n}
\SetSymbolFont{symbols}{bold}{OMS}{cmsy}{b}{n}
\renewcommand\boldmath{\@nomath\boldmath\mathversion{bold}}}
\makeatother
\def\BibTeX{{\rm B\kern-.05em{\sc i\kern-.025em b}\kern-.08em
    T\kern-.1667em\lower.7ex\hbox{E}\kern-.125emX}}

\begin{document}
\history{Received 15 February 2025, accepted 18 March 2025, date of publication 24 March 2025, date of current version 2 April 2025.}
\doi{10.1109/ACCESS.2025.3554138}

\title{\vspace*{-1ex}Enhanced Privacy and Communication Efficiency in Non-IID Federated Learning with Adaptive Quantization and Differential Privacy}
\author{
\uppercase{Emre ARDIÇ}\href{https://orcid.org/0000-0002-4324-8052}{\raisebox{0.7ex}{\includegraphics[height=1.6ex]{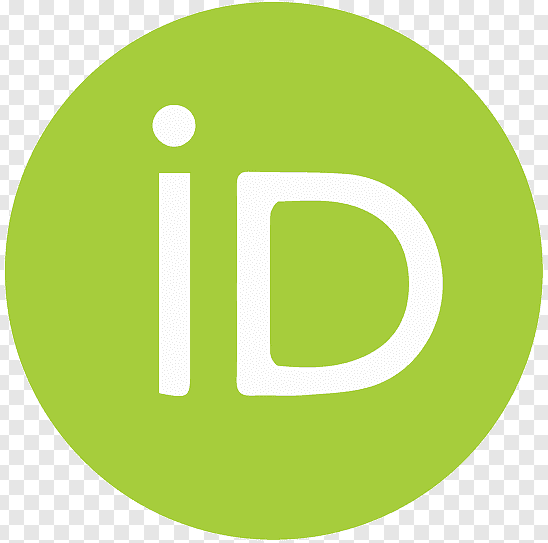}}}
and
\uppercase{Yakup GENÇ}\href{https://orcid.org/0000-0002-6952-6735}{\raisebox{0.7ex}{\includegraphics[height=1.6ex]{orcid.png}}},
(Senior Member, IEEE)
}

\address{Department of Computer Engineering, Gebze Technical University, Gebze, Kocaeli 41400, Türkiye}

\markboth{}{}
\pagestyle{plain}

\corresp{Corresponding author: Emre Ardıç (e-mail: eardic@gtu.edu.tr).}

\begin{abstract}
Federated learning (FL) is a distributed machine learning method where multiple devices collaboratively train a model under the management of a central server without sharing underlying data. One of the key challenges of FL is the communication bottleneck caused by variations in connection speed and bandwidth across devices. Therefore, it is essential to reduce the size of transmitted data during training. Additionally, there is a potential risk of exposing sensitive information through the model or gradient analysis during training. To address both privacy and communication efficiency, we combine differential privacy (DP) and adaptive quantization methods. We use Laplacian-based DP to preserve privacy, which is relatively underexplored in FL and offers tighter privacy guarantees than Gaussian-based DP. We propose a simple and efficient global bit-length scheduler using round-based cosine annealing, along with a client-based scheduler that dynamically adapts based on client contribution estimated through dataset entropy analysis. We evaluate our approach through extensive experiments on CIFAR10, MNIST, and medical imaging datasets, using non-IID data distributions across varying client counts, bit-length schedulers, and privacy budgets. The results show that our adaptive quantization methods reduce total communicated data by up to 52.64\% for MNIST, 45.06\% for CIFAR10, and 31\% to 37\% for medical imaging datasets compared to 32-bit float training while maintaining competitive model accuracy and ensuring robust privacy through differential privacy.
\end{abstract}

\begin{keywords}
Federated learning, adaptive quantization, differential privacy, non-iid distribution
\end{keywords}

\titlepgskip=-21pt

\maketitle

\section{Introduction}
\label{sec:introduction}

\begin{figure}[t]
\begin{center}
    \includegraphics[width=0.77\columnwidth]{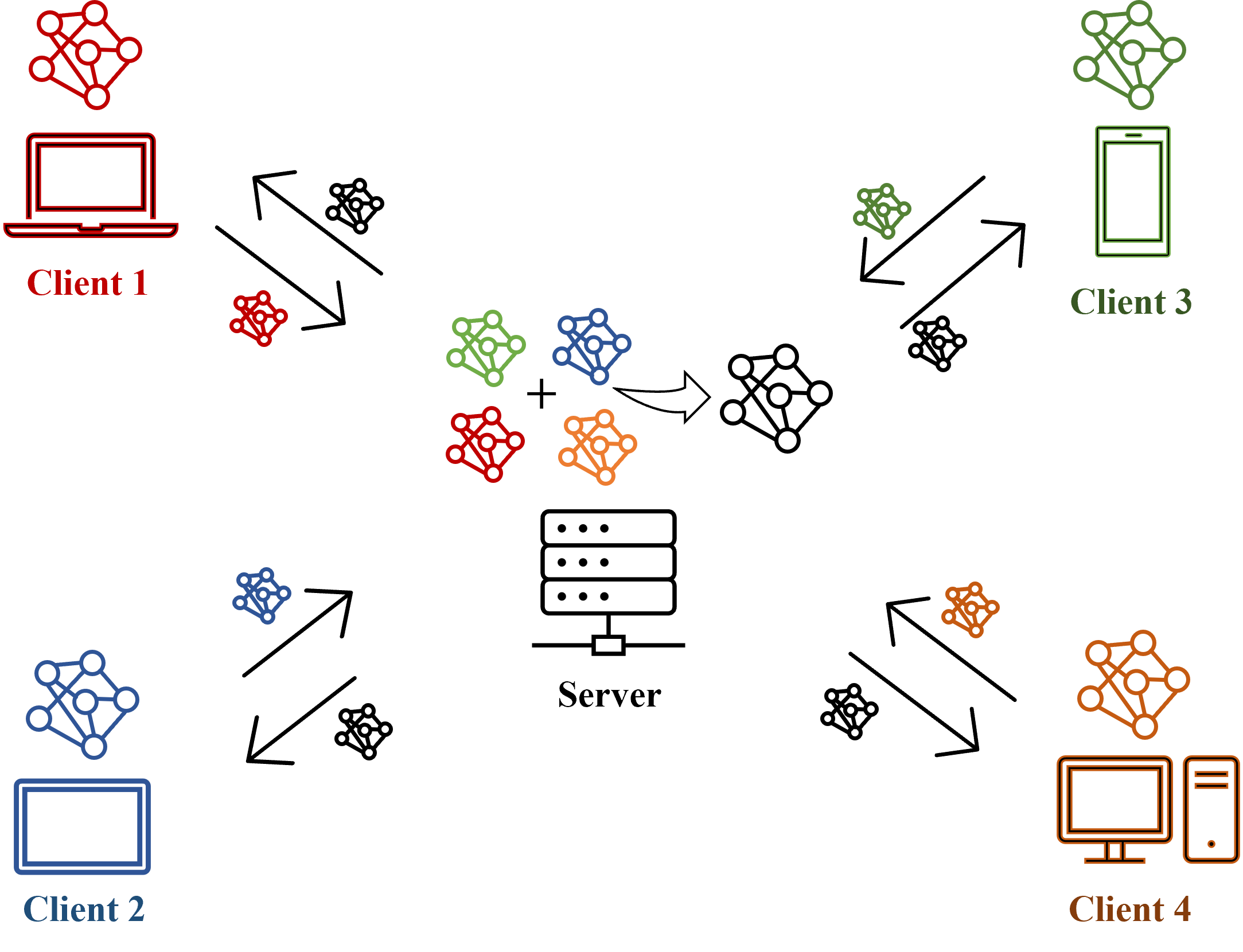}
    \caption{The typical training process of FL with various types of clients and a single server. Best viewed in color.}
    \label{fig:fl_op}
\end{center}
\vspace{-6mm}
\end{figure}

Federated learning (FL) is an emerging distributed machine learning method where multiple devices collaborate to train a model under the coordination of a central server, all without sharing their underlying data \cite{mcmahan2017com}. This approach offers a solution to data privacy concerns inherent in centralized machine learning, where all data is accessible. As the typical learning process is illustrated in Figure \ref{fig:fl_op}, the data of each client is kept local and only model updates are transferred to a global server to optimize a learning objective, thereby increasing data privacy. However, this methodology introduces significant challenges in terms of communication efficiency, privacy, and statistical heterogeneity of data \cite{Li2020}. 

   
The capabilities of devices in an FL network may differ in storage, processing power, battery life, and network connectivity, leading to communication bottlenecks that reduce overall training efficiency. Thus, it's important to optimize communication efficiency by minimizing the total number of communication rounds and the size of transmitted packages. \cite{huang2013depth}. Despite not sharing underlying data, recent studies have shown the possible exposure of sensitive information through gradient analysis during training \cite{bhowmick2018protection, carlini2019secret}. Besides, the size and distribution of data in an FL network may vary widely across individual devices due to their local environment and usage patterns. This non-independent and identically distributed (non-IID) data generation process introduces bias into the training, leading to slow convergence or even divergence \cite{sattler2019robust}. 

To the best of our knowledge, no previous studies have explored the combination of differential privacy and adaptive quantization on non-IID datasets, especially at large scales (e.g., 1000 clients). This gap is crucial as both privacy preservation and communication efficiency are key challenges in real-world FL settings where data is often non-IID and client participation is highly heterogeneous. By addressing these challenges simultaneously, our work provides a novel solution that scales effectively to larger networks of clients, making FL more practical and secure in diverse and distributed environments. Our approach is especially important for applications with stringent privacy requirements and limited communication resources, such as mobile or edge computing environments.

In this work, we propose a novel adaptive quantization approach in which the bit-length is dynamically adjusted using cosine annealing and entropy analysis during training. Specifically, we use different quantization methods for server-to-client and client-to-server communications. For server-to-client transmission, we use a round-based cosine annealing schedule, starting with 32-bit precision and gradually reducing it as training progresses. To optimize client-to-server communication, we employ a combination of round-based cosine annealing and dataset entropy to adjust the bit-length for clients. We compute the Shannon entropy on the local dataset of each client to efficiently measure the amount of information in the dataset \cite{gray2011entropy}. The high entropy indicates that the dataset is balanced and contains multiple classes, thereby leading to increased accuracy and easier convergence of models. The clients with higher entropy tend to have less compression of their weights, thus contributing more to the global model. We validate our approach through extensive experiments on several datasets, including CIFAR10, MNIST, PAP-Smear, Chest X-ray (Pneumonia), and BreakHisV1, showing its effectiveness across varying numbers of clients and privacy budgets in non-IID settings. Our results show significant improvements in communication efficiency while preserving privacy, making a compelling case for adopting our methods in practical FL scenarios. Our key contributions are as follows: 

\renewcommand\labelitemi{\small$\bullet$} 
\begin{itemize} 
\itemsep=-1pt 		
\itemindent=-3pt 	
\item A simple and efficient global bit-length scheduler using round-based cosine annealing to adjust bit-length globally during training, thereby enhancing communication efficiency without compromising model accuracy.
\item A novel adaptive quantization approach using the Shannon entropy to adjust bit-length dynamically on clients during training, thereby prioritizing clients with more information in their datasets.
\item A novel FL framework incorporating Laplacian-based DP and adaptive quantization to enhance both privacy and communication efficiency, extensively evaluated on various model architectures across different numbers of clients, privacy budgets, quantization settings, and non-IID datasets.
\end{itemize}

\section{Related Works}

There exist notable gaps in the literature regarding the integration of adaptive quantization and differential privacy, especially in the context of large-scale FL networks with non-IID data distributions. In this section, we present relevant studies focusing on communication efficiency and privacy, aiming to highlight the intersection of these critical aspects and identify potential areas for further exploration.

\subsection{Communication Efficiency}

Communication bottlenecks can be a big problem due to variations in connection speed and bandwidth across devices in an FL network. It can also be expensive and unreliable if edge devices such as mobile phones have metered connections, and their availability cannot be predicted. To reduce total training time and speed up the convergence of deep learning models in FL settings, communication should be optimized in terms of the total time elapsed and the total size of messages transferred during the training rounds. In addition, using a single global server can create a bottleneck for an FL network with millions of devices. This has led to a significant recent interest in reducing the communication cost of FL, which can be categorized into two groups: decreasing the total number of communication rounds and data compression.

The training process of a typical deep learning model generally takes hundreds of epochs. In vanilla FL, each client sends a local update to a global server after completing its training, which can lead to substantial communication overhead. This overhead can be mitigated by increasing the communication interval, which can be achieved by performing multiple epochs of local training on clients before transmitting updates to the central server. Federated Averaging (FedAvg) and other similar methods use this strategy to reduce the total number of communication rounds \cite{mcmahan2017com}. The communication interval is a tunable hyperparameter that greatly affects the final accuracy of the global model. While short intervals are better for model performance, long intervals increase the convergence speed of the model at the cost of gradient computation. 


There are efficient methods such as sparsification, subsampling, and quantization to reduce the size of messages transmitted at each round \cite{konevcny2016fl, wangni2018gradient, wang2018atomo, caldas1812expanding, tang2018communication, sattler2019robust, lin2017deep, comeff2024, adaptivecomp2024}. Subsampling \cite{konevcny2016fl} and sparsification \cite{wangni2018gradient} methods restrict model updates to a small subset of parameters that are randomly selected or based on predefined criteria. Quantization methods try to reduce the gradients to a smaller set of values. In other words, the original gradients are represented by a low-precision data type such as 1 byte or 1 bit. Since 32-bit is the frequently used data type in deep learning, the maximum compression rate is restricted to 1/32. Convergence speed may also be slower due to information loss, as discussed in \cite{Li2020}. Bernstein et al. propose signSGD to quantize each gradient update to its binary sign, which reduces the size of each gradient value by a factor 32 \cite{bernstein2018signsgd}. Theoretically, it has been shown that this compression method has a convergence guarantee on IID data. Other methods like TernGrad \cite{wen2017terngrad}, QSGD \cite{alistarh2017qsgd}, and ATOMO \cite{wang2018atomo} propose probabilistic approaches to quantize the gradients before sending them to the server. However, one needs to carefully choose the quantization level to balance the learning accuracy and quantization error trade-off. 

Recent advances have focused on adaptive quantization techniques to dynamically adjust quantization precision based on the training stage or model characteristics. For instance, AdaQuantFL \cite{jhun2021adaptive_q} employs an adaptive approach using stochastic uniform quantization to dynamically adjust quantization levels, optimizing the trade-off between communication efficiency and quantization error. Similarly, FedDQ \cite{FedDQ_qu} proposes a descending quantization strategy, speeding up the training convergence by reducing bit-lengths as model updates shrink during training. While effective, this method assumes uniform client contributions and overlooks dataset diversity and heterogeneity. More recently, FedAQ \cite{qu2024fedaq} proposes a joint uplink and downlink adaptive quantization strategy that optimizes communication for resource-constrained environments, but it does not address non-IID data distributions or privacy concerns. To overcome these limitations, we propose a descending quantization strategy for both uplink and downlink communications, using a novel client importance estimation mechanism that is designed to handle non-IID datasets effectively. In addition, unlike prior studies, we rigorously evaluate our method on a large scale, with experiments involving up to 1000 clients. This design addresses the critical gaps in downlink optimization and data heterogeneity found in existing approaches while significantly improving communication efficiency and scalability.

\subsection{Privacy} \label{sec_privacy}

The main motivation is to make sure that raw data on each client remains local in an FL network. It has been shown that the gradients or model updates can reveal sensitive information about the data \cite{bhowmick2018protection, carlini2019secret}. For example, Carlini et al. show that sensitive text data such as credit card numbers can be revealed by analyzing recurrent neural network models trained on the language data of users \cite{carlini2019secret}. In the literature, recent works use differential privacy, secure multiparty computation (SMC), and other cryptographic algorithms to preserve the privacy of transmitted messages and local data on each client. In FL, low device participation tolerance, computation, and communication efficiency are necessary without significantly decreasing model accuracy. 

Differential privacy is a frequently used randomized method to reduce the relation between the input and output of deep learning models. Ideally, a change in one input feature is expected not to cause too much difference in the output distribution. The aim is to hide whether or not a specific sample is used in the learning process. This sample-level privacy can be used in many machine learning applications \cite{konevcny2016fl, abadi2016deep,iyengar2019towards}. For gradient-based learning, a simple and popular approach is the random perturbation of the outputs at each iteration \cite{iyengar2019towards}. The gradients can be clipped to restrict the effect of each sample on the entire update. The perturbation can be done by adding noise with probabilistic methods such as Laplacian \cite{zhou2023exploring}, Gaussian \cite{abadi2016deep} or Binomial \cite{melis2015efficient}. Since increasing the perturbation rate can decrease the accuracy of the model, it can be hard to balance this trade-off.

Recent studies on privacy in FL can be split into two categories: global and local privacy \cite{carlini2019secret, geyer2017dpfl, li2019dp, mcmahan2017learning, andrew2021dp, bonawitz2017practical, fu2024dpfl, tfl2024}. In global privacy, the model updates computed during training iterations are private to all untrusted third parties except the global server, whereas local privacy means that the updates are also private to the server. Cryptographic protocols like SMC, which are lossless and highly secure, can improve the privacy of FL without compromising the learning accuracy, as demonstrated by Bonawitz et al. \cite{bonawitz2017practical}. However, these protocols often incur high communication costs.

Several studies have focused on methods to address both privacy and communication efficiency in FL. Lang et al. proposed JoPEQ that combines lossy compression with privacy enhancement through vector quantization and local DP \cite{jopeq_lang}. However, it employs static quantization rather than adaptive quantization strategies tailored to client or parameter importance. Similarly, MSPDQ-FL uses a model-splitting strategy with dynamic quantization to enhance privacy and reduce communication costs in non-IID data settings \cite{mspdqfl_wang2024}. However, its dynamic quantization is focused on submodel parameters rather than the full parameter space. Youn et al. introduce the Randomized Quantization Mechanism (RQM) that achieves privacy by combining randomized sub-sampling and rounding of quantization levels, satisfying Renyi DP \cite{rqm_youn2023}. Despite its innovative approach to privacy-preserving quantization, RQM primarily focuses on privacy-accuracy trade-offs and does not incorporate adaptive mechanisms. Finally, Nguyen et al. propose a framework integrating quantization and Binomial noise to optimize privacy and communication parameters, yet it lacks adaptive quantization mechanisms and explicit scalability analysis for large-scale FL setups \cite{optfl_nguyen2023}. 

In this work, we employ Laplacian-based DP instead of Gaussian-based DP due to its sharper privacy guarantees under \( \ell_1 \)-sensitivity, making it ideal for bounded updates in FL \cite{zhou2023exploring}. We first apply Laplace noise to secure model updates, followed by adaptive quantization that dynamically reduces bit-lengths based on a novel client importance estimation mechanism. This consecutive usage ensures that the added noise is effectively incorporated into the quantized updates, maintaining a balance between privacy, accuracy, and communication cost, particularly in non-IID and large-scale FL settings.

\begin{table}[t]
\centering
	\caption{Notations.}
        \label{table:notations}
	\begin{tabular}{ll}
		\toprule
            Symbol & Description  \\ 
		\midrule
            $\mathcal{N}$ & the set of all clients \\
            $N$         & the total number of clients \\
            $\mathcal{P}_t$ & the set of selected clients in the $t$-th round \\
            $P$ & the number of selected clients per round \\
            $T$         & the total communication rounds \\
            $t$         & the round index \\
            $E$         & the number of local training epochs  \\
            $B$         & the batch size  \\
            $\zeta_i$   & the client $i$\\
            $D_i$       & the dataset of client $i$\\
            $K$         & the total number of classes \\
            $n_i$       & the number of samples owned by client $i$  \\
            $n^t_{max}$ & the biggest dataset size among clients in the $t$-th round \\
            $\ell$ & the loss function \\
            $\mathcal{L}_i$ & the overall loss at client $i$\\
            $\eta$ & the learning rate \\
            $Q$         & the quantization function \\
            $\mathcal{S}^i_t$  & the scale factors computed on client $i$ in the $t$-th round \\
            $DQ$        & the dequantization function \\
            $b$         & the quantization bit-length \\
            $\nu_i$     & the importance score of client $i$ \\
            $\lambda_h$ & the weight of dataset homogeneity score for $\nu_i$ \\
            $\rho$      & the stochastic rounding function \\
            $\theta_t$  & the model parameters in the $t$-th round \\
            $\xi$       & the gradient norm value \\
            $\epsilon, \delta$  & the DP parameters \\
            $\Xi^i_t$   & the sensitivity of client $i$ in the $t$-th round \\
            $\nabla f$  & the gradient of the function $f$ \\
            $\lambda_i$ & the local Lipschitz smoothness on client $i$ \\
            $w^i_t$     & the noise generated by DP on client $i$ in the $t$-th round \\
	      \bottomrule
	\end{tabular}
\end{table}

\section{Proposed Methods} \label{sec_methods}

In this section, we begin by clearly defining FL and providing a detailed step-by-step explanation of the training process. Next, we introduce our training datasets, which are distributed to clients in a non-iid manner, and detail the classification models optimized for FL. Then, we introduce our DP approach, which utilizes Laplacian noise. Finally, we discuss our adaptive quantization methods, which use Shannon entropy and cosine annealing to enhance communication efficiency between the server and clients.

\subsection{Federated Learning}

We consider a typical synchronous FL strategy with two types of elements: a server and $N$ clients $\mathcal{N}=\{\zeta_1,\cdots, \zeta_N\}$, where each client $\zeta_i$ has a local dataset $D_i=\{(x_j,y_j)\}_{j=1}^{n}$, with $x_j$ representing a sample and $y_j\in[c_1,..,c_K]$ representing a label. The main objective is to train a single global model by using a large number of clients. In this process, only the model parameters are periodically transferred between the server and clients. We use the FedAvg algorithm to aggregate local models collected from remote devices in each iteration \cite{mcmahan2017com}. As shown in Figure~\ref{fig:fl_op}, a central server manages the training process over iterations by repeating the following steps:

\begin{enumerate}
\item \textit{Client Selection:} The central server chooses active clients based on criteria like battery level, network quality, and CPU load to minimize any adverse effects on device performance and usability.
\item \textit{Broadcasting:} The chosen clients retrieve the current model weights and training parameters from the central server.
\item \textit{Local Update:} Each selected client performs a model update using its local data.
\item \textit{Aggregation:} The server gathers and combines the updates from all clients. During this phase, quantization and DP methods can be applied to lower communication costs and ensure data privacy.
\item \textit{Global Update:} The server updates the global model based on the aggregated updates received from the selected clients in the current round.
\end{enumerate}

Let $\ell(f(x_j; \theta), y_j)$ denote the loss function for a sample $(x_j, y_j) \in D_i$ in client $\zeta_i$, where $x_j$ is an input sample and $y_j$ is a ground-truth label. The model can be defined as $f(x_j; \theta)$ where $\theta$ is the model parameters. The overall loss $\mathcal{L}_i(\theta)$ at a client $\zeta_i$ is defined in Eq. \eqref{eq_fiw} where $n_i=|D_i|$ represents the total number of samples. As the training progresses, the main objective of each client is to minimize the loss function $\mathcal{L}_i(\theta)$ on its local dataset.
\begin{equation}
\mathcal{L}_i(\theta) = \frac{1}{n_i}\sum_{(x_j, y_j \in D_i)}\ell(f(x_j; \theta), y_j)
\label{eq_fiw}
\end{equation}

In FL, the main purpose is to minimize the objective function as defined in Eq. \eqref{eq_fl} \cite{Li2020}. This objective function is solved in a distributed manner where each client conducts multiple epochs of training on its local model using a stochastic gradient descent (SGD) optimizer. In this study, we use the Cross-Entropy (CE) loss function during local training on clients. 
\begin{equation}
\hat{\theta}=\arg\min_\theta\frac{\sum_{i=1}^{N}n_i\mathcal{L}_i(\theta)}{\sum_{i=1}^{N}n_i}
\label{eq_fl}
\end{equation}

\subsection{Datasets} \label{sec_methods_dataset}

In this study, we train and test our FL models by using MNIST \cite{deng2012mnist} and CIFAR10 \cite{cifar10} datasets. The MNIST dataset, a classic benchmark in machine learning, comprises grayscale images sized 28x28 pixels. These images represent handwritten digits ranging from 0 to 9 and are sourced from a diverse group of 1,000 individuals. To simulate a diverse setting, we distribute the data across 1000 clients, with each client holding samples of only two different digits. The number of samples per client follows a power law distribution. This dataset is aligned with the study of Li et al. \cite{li2020fl_mnist}.

On the other hand, the CIFAR10 dataset is another well-established benchmark, but it features color images with dimensions of 32x32 pixels. This dataset contains images of 10 distinct categories, including vehicles (such as cars and trucks) and animals (like birds and cats). Each image is labeled according to its category, providing a more complex challenge compared to MNIST due to the variability in object appearances and backgrounds. The detailed statistics of the datasets are shown in Table \ref{table:datasets}.

\begin{table}[h]
    \centering
	\caption{The details of the commonly used CIFAR10 and MNIST datasets and related models used in this study.}
        \label{table:datasets}
	\begin{tabular}{lll}
		\toprule
            Dataset & CIFAR-10 & MNIST \\
		\midrule
            Image Size          & 32x32 RGB & 28x28 Gray \\
            Train Samples       & 50,000    & 61,664  \\
            Test Samples        & 10,000    & 7,371  \\
            \# Classes          & 10        & 10 \\
            Model               & VGG7      & 2-Layered CNN \\
            \# Parameters       & 300K      & 1.6M \\
		\bottomrule
	\end{tabular}
\end{table}

We use a Dirichlet distribution with an alpha parameter of 0.5 to distribute the CIFAR10 dataset among clients, thereby simulating a realistic non-IID scenario. This approach introduces a greater degree of data heterogeneity, with certain clients receiving a disproportionate share of specific classes. This mimics real-world data distributions, which are often uneven across different sources. For the MNIST dataset, which comprises handwritten digits from 1,000 users, we assign each user as a client in our FL network, naturally creating a non-IID partitioning. When working with fewer than 1,000 clients, we randomly combine images from multiple users to meet the required number of clients.

\subsection{Model Architectures} \label{sec_model_archs}

\begin{figure}[t]
    \centering
    \includegraphics[width=1.0\columnwidth]{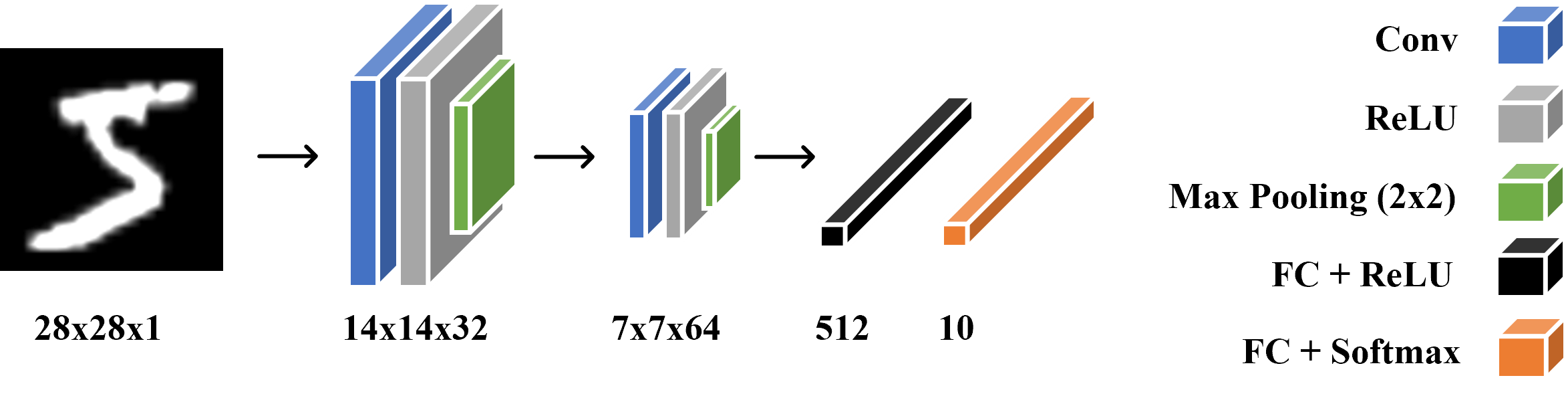}
    \caption{The CNN architecture designed for MNIST dataset. Best viewed in color.}
    \label{fig:model_mnist}
\end{figure}

The model architecture used for the MNIST dataset is presented in Fig.~\ref{fig:model_mnist}. We employ a simple two-layered convolutional neural network (CNN) consisting of 32 and 64 filters, which is also used in the FedAvg study \cite{mcmahan2017com}. The kernel size, stride, and padding parameters of the CNN layers are set to 5, 1, and 2. Following each convolution layer, we use the Rectified Linear Unit (ReLU) activation function and a max-pooling layer with a kernel size of 2x2 and a stride of 2. The output from the final convolutional layer is then flattened and passed through a fully connected (FC) layer with ReLU, producing a 1D output of length 512. The final classification is performed by an FC layer with a Softmax activation function, which maps the output to one of the ten classes.

\begin{figure}[t]
    \centering
    \includegraphics[width=1.0\columnwidth]{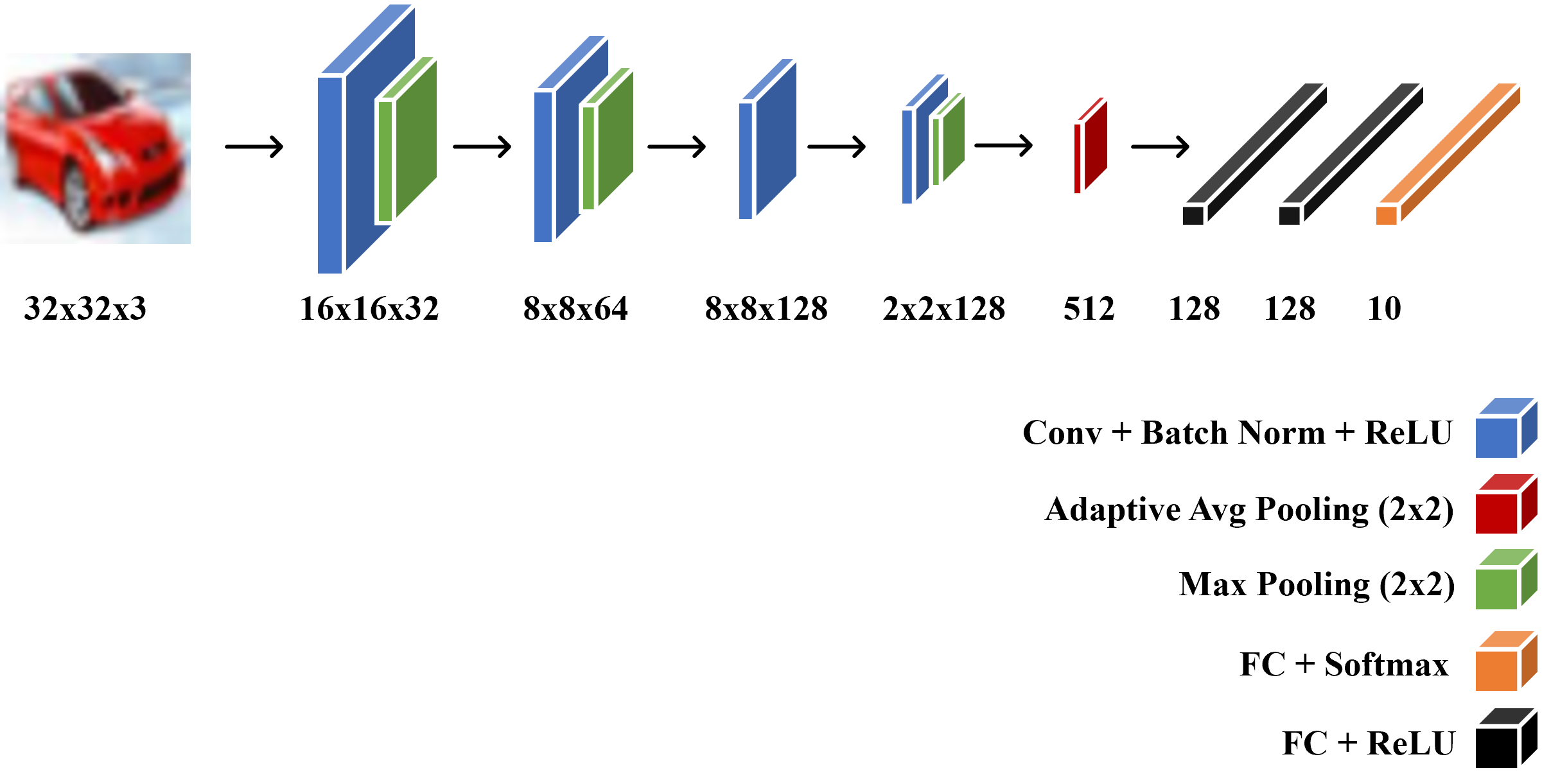}
    \caption{The VGG7 architecture designed for CIFAR10 dataset. Best viewed in color.}
    \label{fig:model_cifar}
\end{figure}

The architecture of the model for the CIFAR10 dataset with 32x32 color images is presented in Fig.~\ref{fig:model_cifar}. The model consists of a four-layered CNN consisting of 32, 64, 128, and 128 filters, respectively. Each convolutional layer employs a 3x3 kernel, a stride of 1, and a padding of 1. After each convolutional layer, we apply the ReLU function and batch normalization to ensure stable and efficient training. The output from the final convolutional layer is passed through adaptive average pooling with a window size of 2x2, resulting in an output size of 2x2x128. The output is then flattened into a 1D vector of length 512. Following this, two FC layers with the ReLU function are used, producing a 1D vector of length 128. The final classification layer is an FC layer with a Softmax function that is responsible for predicting the ten classes.

\subsection{Stochastic Uniform Quantization} \label{sec_quant}

In this work, we employ the stochastic uniform quantization to reduce the size of the model weights by mapping them to a smaller set of values, as used in these studies \cite{alistarh2017qsgd, jhun2021adaptive_q, FedDQ_qu}. This type of quantization is used to map continuous values to discrete levels with an element of randomness, ensuring that the quantization error is uniformly distributed. This leads to a more robust representation, especially in machine learning applications where precision and generalization are critical. The quantization process involves two fundamental operations as follows:

\renewcommand\labelitemi{\small$\bullet$} 
\begin{itemize} 
\itemsep=-1pt 		
\itemindent=-3pt 	
\item Quantize operation converts a real number to a quantized integer representation (e.g., from FP32 to INT8).
\item Dequantize operation converts a quantized integer representation to a real number (e.g., from INT8 to FP32).
\end{itemize}

Let $[\beta, \alpha]$ be the range of representable real values, and $b$ be the target bit-width of the signed integer representation. Uniform quantization transforms the input value $x \in [\beta, \alpha]$ to the range $[-2^{(b-1)}, 2^{(b-1)}-1]$, where inputs outside the range are clipped to the nearest bound. Considering only uniform transformations, there are only two options for the transformation function: $g(x) = s \cdot x + z$ and its special case $g(x) = s \cdot x$, where $x, s, z \in \mathbb{R}$. These two choices are referred to as affine and symmetric transformations, respectively \cite{wu2020int_q}. 


In this work, we employ symmetric quantization ($z=0$) with stochastic rounding that performs range mapping with only a scale transformation where the input range and integer range are symmetric around zero. This type of quantization is defined in Eq.~\eqref{eq:quant_q} where $\rho$ is the stochastic rounding operation, $s$ is the scale factor, and $clip$ is the clipping function. The scale factor $s$ is computed by Eq.~\eqref{eq:quant_s} where $b$ is the target bit-length and $\alpha \in \mathbb{R}$ is the maximum real value in our tensor with arbitrary dimension.
\begin{equation}
Q(x,s,b)=clip(\rho(x * s), b)
\label{eq:quant_q}
\end{equation}
\begin{equation}
s = \frac{2^{(b-1)} - 1}{\alpha}
\label{eq:quant_s}
\end{equation}

The clip operation $clip(y, b)$ is defined in Eq.~\eqref{eq:quant_clip} where $y \in \mathbb{Z}$ is an integer value.
\begin{equation}
clip(y, b) = \begin{cases} 
    -2^{(b-1)}+1 & \text{if } y < -2^{(b-1)}+1 \\
    2^{(b-1)}-1 & \text{if } y > 2^{(b-1)}-1 \\
    y & \text{otherwise}
\end{cases}
\label{eq:quant_clip}
\end{equation}

The stochastic rounding operation $\rho:\mathbb{R} \to \mathbb{Z}$ is defined in Eq.~\eqref{eq:quant_rho}, which is a method used to convert a floating-point number to a fixed-point or integer representation, where the number is rounded to one of its nearest values with a probability that is proportional to its distance from those values. 
\begin{equation}
\rho(x) = 
\begin{cases} 
\lfloor x \rfloor & \text{with probability } p = \lceil x \rceil - x \\
\lceil x \rceil & \text{with probability } 1 - p = x - \lfloor x \rfloor
\end{cases}
\label{eq:quant_rho}
\end{equation}

The dequantize operation, defined in Eq.~\eqref{eq:quant_deq}, converts a number from a quantized integer representation $x_q \in \mathbb{Z}$ to a real number. The scale is computed in the quantization phase as described in Eq.~\eqref{eq:quant_s}.
\begin{equation}
DQ(x_q, s) = \frac{x_q}{s}
\label{eq:quant_deq}
\end{equation}


It is possible to share quantization parameters among tensor elements, a concept known as quantization granularity \cite{wu2020int_q}. The finest granularity uses individual quantization parameters per element, which can be computationally expensive for large tensors. In this work, we employ per-tensor granularity, sharing the same quantization parameters for each three-dimensional tensor in each layer of the model.

\subsection{Adaptive Quantization} \label{sec_quant}

We propose two novel adaptive quantization methods to adjust the quantization precision dynamically throughout the training process. Specifically, we use a cosine annealing schedule to gradually reduce the bit-length allocated for quantization, allowing the model to train with higher precision $b_{max}$ and progressively move towards lower precision $b_{min}$ during the training. This method helps mitigate potential loss in accuracy due to quantization by smoothing the transition across different bit-lengths. We use this descending scheduling concept, which is suggested by Qu et al. \cite{FedDQ_qu} for the following reasons:
\begin{itemize}
\item A high quantization level in the early training stages accelerates loss reduction, but a low level saves bit volume at the cost of slower convergence.
\item  As the model stabilizes later in training, fewer bits are needed to represent updates, making descending-trend quantization more efficient for FL.
\end{itemize}

\begin{figure}[t]
    \centering
    \includegraphics[width=0.7\columnwidth]{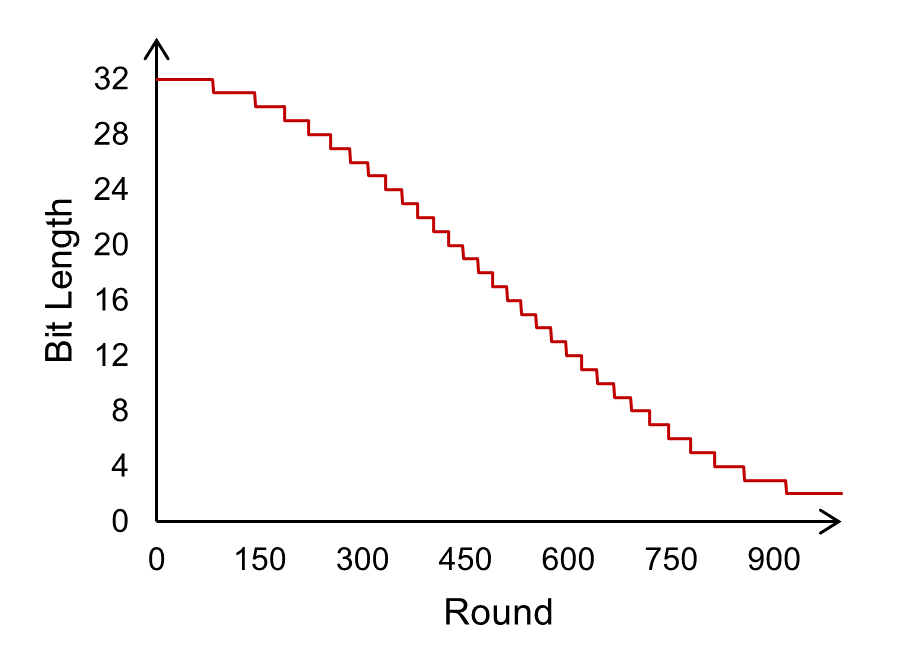}
    \caption{The bit-length scheduling, guided by cosine annealing, begins at 32 bits and gradually reduces to 2 bits.}
    \label{fig:q_cos_sch}
\end{figure}

The bit-length scheduling formulation for a client $i$ in the $t$-th round is defined in Eq.~\eqref{eq:cos_ann}, where $T$ is the total communication rounds, $\nu_i \in [0,1]$ is the importance score of a client $i$, $b_{max}$ is the initial bit-length, and $b_{min}$ is the final bit-length during training. Our first approach uses uniform bit-length scheduling for all clients, gradually reducing the bit-length from $b_{max}$ to $b_{min}$ using cosine annealing as the training progresses. In this approach, we assume $\nu_i=1$ in Eq.~\eqref{eq:cos_ann} for the server and all clients. The cosine annealing scheduling process from 32 bits to 2 bits is demonstrated in Fig.~\ref{fig:q_cos_sch}.
\begin{equation}
b^i_t = b_{min} + \nu_i (b_{max} - b_{min}) \frac{1 + \cos\left(\frac{\pi t}{T}\right)}{2}
\label{eq:cos_ann}
\end{equation}

In our second approach, we introduce a combined strategy of cosine annealing with client-based Shannon entropy for client-to-server transmissions while using only cosine annealing for server-to-client transmissions, as defined in Eq.~\eqref{eq:cos_ann}. In this hybrid approach, the cosine annealing schedule determines the overall trend in bit-length reduction, while adjustments based on client-based entropy ensure that the quantization process adapts to each client's specific data distribution and characteristics. The main idea is to allocate higher bit-lengths to clients with larger datasets that contain many classes and exhibit balanced distributions, thereby accelerating model convergence and potentially improving accuracy. To achieve this, we propose the formulation in Eq.~\eqref{eq:shannon_ent} to estimate client importance score $\nu_i$, where $p^i_k$ represents the probability of a class $k$, $\lambda_{h} \in [0,1]$ represents the weight of dataset homogeneity score, and $n^t_{max}$ is the number of samples of the client with the most samples in the $t$-th round, computed as $n^t_{max}=\max_{i \in P_t} |D_i|$. 

The client importance $\nu_i$ in Eq.~\eqref{eq:shannon_ent} assigns higher scores to clients with larger and more homogeneous datasets, indicating their greater role in model convergence. The importance of dataset size is computed by $\nicefrac{|D_i|}{n^t_{max}}$, prioritizing clients with larger datasets in the $t$-th round, as they provide more statistically significant updates. Meanwhile, the dataset homogeneity score of a client is estimated using normalized Shannon entropy, ensuring that clients with more diverse and balanced class distributions receive higher entropy scores while those with concentrated class distributions have lower scores. Both scores are designed to fall within the range \([0,1]\) and are balanced using the weighting factor \( \lambda_h \), which controls the trade-off between dataset size and homogeneity in determining client importance. 

\begin{equation}
\nu_i = \lambda_{h} \frac{-\sum_{k=1}^{K} p^i_k \log_2(p^i_k)}{\log_2(K)} + (1-\lambda_{h}) \frac{|D_i|}{n^t_{max}}
\label{eq:shannon_ent}
\end{equation}

\subsection{Differential Privacy} \label{sec_dp}

As discussed in Section~\ref{sec_privacy}, although FL addresses some privacy concerns, it does not fully prevent leakage of sensitive information through model updates \cite{wei2020federated, ma2020safeguarding, wang2019beyond}. Hence, additional methods like DP are necessary to mitigate such risks. However, although DP achieves privacy by injecting artificial noise, the resulting increase in noise often comes at the expense of reduced model accuracy. To avoid information leakage, each client should locally perturb its trained parameters by deliberately introducing noise before uploading them for aggregation at the server.

A randomized function $M:X \rightarrow R$ defined over the domain $X$ and the range $R$ is $(\epsilon, \delta)$-differentially private if for any pair of neighboring datasets $D, D^{\prime} \in X$ differing by one record, and for any $S\subseteq R$, the condition in Eq.~\eqref{eq:dp} is satisfied. Here, $\epsilon>0$ denotes the privacy budget, and $\delta \geq 0$ represents the probability of privacy leak. A larger $\epsilon$ with a fixed $\delta \leq 1$ provides a more distinct differentiation between these neighboring datasets, consequently leading to an increased risk of privacy violation. Thus, the choice of $\epsilon$ causes a trade-off between accuracy and privacy.
\begin{equation}
\Pr[M(D) \in S] \leq e^{\epsilon} \Pr[M(D^{\prime}) \in S] + \delta
\label{eq:dp}
\end{equation}

The use of DP can be classified as Central DP, Local DP, or a combination of both. Local DP uses DP on clients before sending model weights to the server, while Central DP uses it at the server before broadcasting the global model to clients. In this study, we employ Local DP with $\delta=0$ during the training process.
\begin{equation}
\Xi = \max_{\forall \theta, \mathcal{D}': \|\mathcal{D} - \mathcal{D}'\|_1 = 1} \| \Phi(\theta, \mathcal{D}) - \Phi(\theta, \mathcal{D}') \|_1
\label{eq:dp_l1sens}
\end{equation}

During local training, we apply gradient clipping to bound the $l_1$-norm of the gradients of the loss function $\ell$, which is important for sensitivity analysis, as discussed in \cite{zhou2023exploring}. Considering each client has statistically different datasets due to non-IID distribution, the $L_1$-sensitivity of client $\zeta_i$ in the $t$-th round, denoted by $\Xi^i_t$, is determined by Eq.~\eqref{eq:dp_sens}, where $\xi$ is the $l_1$-norm gradient clipping, $\eta$ is the learning rate, $E$ is the number of local epochs, $n_i=|D_i|$ is the total samples, and $E_0$ is the minimum integer such that $(1 + \lambda_i \eta)^{E_0} \geq 1 + n_i$.
\begin{equation}
\Xi_t^i = \begin{cases} 
\frac{2 \xi E \eta}{n_i}, & \text{if } \lambda_i = 0, \\
\frac{2 \xi}{\lambda_i n_i} \left( (1 + \lambda_i \eta)^E - 1 \right), & \text{if } \lambda_i > 0, E < E_0, \\
2 \xi + 2 \eta \xi (E - E_0), & \text{if } \lambda_i > 0, E \geq E_0.
\end{cases}
\label{eq:dp_sens}
\end{equation}

In this mechanism, noise $w^i_t$ is added to the model weights of the client $\zeta_i$ in the $t$-th round before transmitting the updated weights to the central server. The noise is computed by $w^i_t=Lap(0,\frac{T_i}{\epsilon}\Xi^i_t)$, where $Lap(\mu, s)$ represents the Laplace distribution with mean $\mu$ and scale $s$ of the same dimension as $w^i_t$. The sensitivity is scaled by $T_i=\frac{PT}{NE}$, where $P$ is the number of clients that participated in the $t$-th round, $T$ is the total communication round, and $N$ is the total clients. 

The noise level on each client is adjusted by the sensitivity that is mainly influenced by the number of samples and a regularization parameter $\lambda_i$. Here, $\lambda_i$ is the local Lipschitz smoothness of the loss function $\ell$ on client $\zeta_i$, which is used to limit the gradient norms. This is important for controlling the sensitivity of the local updates and helps in reducing the noise magnitude required by the DP mechanisms, thereby improving the model's utility while preserving privacy.

To align with the $L_1$-sensitivity of Laplace noise, we use $l_1$-norm to compute $\lambda_i$ as defined in Eq.~\eqref{eq:dp_lambda}, where $\nabla f(\theta)$ represents the gradient of the function $f$ evaluated at the parameter $\theta$. Here, note that $f$ is the learning model, $\theta$ is the initial model parameter, and $\theta'$ is the final model parameter after local training.
\begin{equation}
\lambda_i = \frac{\|\nabla f(\theta) - \nabla f(\theta')\|_1}{\|\theta - \theta'\|_1}
\label{eq:dp_lambda}
\end{equation}

Since we perform mini-batch training with multiple local epochs, we track the gradient and parameter differences for each batch across successive epochs on clients. Using these differences, we compute $\lambda_i$ values for each batch and use the largest value among all $\lambda_i$ values as the final $\lambda_i$. 

\begin{algorithm}[t]
\small
\caption{The proposed FedAvg algorithm combined with adaptive quantization and differential privacy.}
\label{alg:client_dpq_fedavg}
\begin{algorithmic}[1]

\Function{Server}{$T, \mathcal{N}, b_{max}, b_{min}, E, B$}
    \State initialize $\theta_0$
    \For{$t = 0$ to $T - 1$}
        \State select $P$ clients randomly as $\mathcal{P}_t \subset \mathcal{N}$
        \State $b_t = b_{min} + (b_{max} - b_{min}) \frac{1 + \cos\left(\frac{\pi t}{T-1}\right)}{2}$
        \State $\hat{\theta}_t, \mathcal{S}_t \gets Quantize(\theta_t, b_t)$
        \State $n^t_{max}=\max_{\zeta_i \in \mathcal{P}_t} |D_i|$
        \State $n=0$
        \For{each client $\zeta_i \in \mathcal{P}_t$ \textbf{in parallel}}
            \State $\hat{\theta}^{i}_{t}, n_i, \mathcal{S}^i_t \gets Client(\hat{\theta}_t, n^t_{max}, \mathcal{S}_t, E, B)$
            \State $\theta^{i}_{t} \gets Dequantize(\hat{\theta}^{i}_{t}, \mathcal{S}^i_t)$
            \State $n = n + n_i$
        \EndFor
        \State $\theta_{t+1} \gets \sum_{\zeta_i \in \mathcal{P}_t} \cfrac{n_i \theta^{i}_{t}}{n}$
    \EndFor
\EndFunction
\Statex
\Function{Client}{$\hat{\theta}_{t}, n^t_{max}, \mathcal{S}_t $}
    \State $\theta_t^{i} \gets Dequantize(\hat{\theta}_{t}, \mathcal{S}_t)$
    \For{$e = 0$ to $E-1$}
         \For{each batch of size $B$ in the dataset $D_i$}
            \State $\theta_{t}^{i} \gets \theta_t^{i} - \eta \nabla f(\theta_t^{i}) / \max(\|\nabla f(\theta_t^{i})\|_1, \xi)$
            \State \textit{where $\xi$ is the clipping bound.}
          \EndFor
    \EndFor
    \State generate noise for DP: $w_t^i \gets \text{Lap}\left(0, \frac{T_i}{\epsilon} \Xi^i_t \right)$
    \State $\theta_{t}^{i} \gets \theta_{t}^{i} + w_t^{i}$    
    \State compute client utility score $\nu_i$ by Eq.~\eqref{eq:shannon_ent}
    \State compute quantization bit-length $b^i_t$ by Eq.~\eqref{eq:cos_ann}
    \State $\hat{\theta}_t, \mathcal{S}^i_t \gets Quantize(\theta_t, b^i_t)$
    \State \Return $\hat{\theta}_t, \mathcal{S}^i_t$
\EndFunction
\end{algorithmic}
\end{algorithm}

\section{Experiments}

In this section, we present a comprehensive evaluation of our FL approach defined in Algorithm~\ref{alg:client_dpq_fedavg}, across various model architectures, client counts, privacy budgets, and non-IID datasets. Our primary goal is to show that our method not only converges with significantly fewer communicated bits but also maintains model accuracy and enhances privacy. We first evaluate our DP and quantization methods individually, then assess their combined impact under various client counts and privacy budgets. Throughout the training process, we monitor the number of rounds that yield the highest test accuracy, computed on the server using the global model, referred to as the "Best Round" in experiments.

\subsection{Deployment} \label{sec_exp_deploy}

To evaluate our methods in an FL environment, we employ the FedML library, which supports single-machine simulations, distributed computation, and edge device training, alongside a versatile programming interface with baseline implementations for optimizers, models, and datasets \cite{fedml2020}. Leveraging this library, we establish an FL system with varying client counts on a single machine with NVIDIA RTX 3090 GPU, Ryzen 5900X CPU, 32GB RAM, and 1TB SSD by using Python 3.6, Scikit-Learn 0.24.2, PyTorch 1.8.2, and CUDA 11.1. We extend this library by integrating DP and adaptive quantization methods as well as our models detailed in Section~\ref{sec_model_archs}. Additionally, We customize the single-process simulation module of FedML for our specific experiments.

\subsection{Training Configuration} \label{sec_train_conf}

The models described in Section~\ref{sec_model_archs} are trained using the FedAvg algorithm for 1000 rounds with varying numbers of clients. Every 10 rounds, we calculate the train and test accuracies on the central server. During training, we record the rounds with the highest test accuracy. We use a Stochastic Gradient Descent (SGD) optimizer with a learning rate of 0.1 and weight decay of 0.001. The batch size is 64, and each client performs five epochs of local training. Our experiments involve 1000, 200, 100, and 50 client numbers, with 100, 20, 10, and 5 clients selected per round, respectively. We maintain a 0.1 ratio between the number of clients per round and the total number of clients. We use the same random seed in all training sessions to ensure a fair evaluation.

\subsection{Laplacian-based DP} \label{sec_exp_dp}

In this section, we extensively evaluate our DP approach detailed in Section~\ref{sec_dp} across varying client counts and privacy budgets $\epsilon$. We first analyze the impact of gradient norm on model accuracy, then examine how varying privacy budget $\epsilon$ affects the accuracy across various client counts on non-IID datasets. 

Gradient clipping is essential for ensuring DP during model training as it limits the influence of individual data points by bounding the sensitivity of gradients \cite{andrew2021dp}. By capping the gradient norms, we prevent any single data point from disproportionately affecting the model, allowing for controlled noise addition. To find the optimal gradient norm $\xi$, we conduct experiments with 100 clients across varying norms. The results presented in Table~\ref{table:exp_by_gnorms} prove that the accuracy diminishes as the norm value $\xi$ is reduced. For example, a norm of 100 leads to accuracy drops of 6.67\% for the CIFAR-10 dataset and 5.22\% for the MNIST dataset, compared to an unbounded norm ($\xi=\infty$). This is expected because gradient clipping limits the magnitude of updates, restricting the model's learning capacity. As described in Eq.~\eqref{eq:dp_sens}, increasing $\xi$ generates more noise, enhancing privacy but also obscuring the true gradient signals, leading to less precise model updates and impaired learning. Therefore, we use $\xi=100$ in subsequent experiments to balance privacy and model stability.

\begin{table}[h]
\centering
    \caption{The best accuracies for the MNIST and CIFAR10 datasets for 100 clients across varying gradient clipping norms.}
        \label{table:exp_by_gnorms}
    \begin{tabular}{lllll}
        \toprule
        & \multicolumn{2}{c}{MNIST} & \multicolumn{2}{c}{CIFAR10} \\ 
        \midrule
        Gradient Norm $\xi$ & Accuracy & Round & Accuracy & Round  \\ 
        \midrule        
        10       & 80.95 & 810 & 61.05 & 930 \\
        20       & 82.49 & 810 & 67.86 & 930 \\ 
        50       & 87.38 & 960 & 73.05 & 930 \\ 
        100      & 93.71 & 970 & 75.30 & 930 \\
        $\infty$ & 98.93 & 830 & 81.97 & 930 \\
        \bottomrule
    \end{tabular}
\end{table}
\begin{table}[h]
\centering
    \caption{The best accuracies for the MNIST and CIFAR10 datasets for a gradient norm of $\xi=100$ across varying client counts.}
        \label{table:exp_by_clients}
    \begin{tabular}{lllll}
        \toprule
        & \multicolumn{2}{c}{MNIST} & \multicolumn{2}{c}{CIFAR10} \\ 
        \midrule
        \# Clients & Accuracy & Round & Accuracy & Round \\
        \midrule
        50     & 95.06 & 850 & 76.48 & 900 \\
        100    & 93.71 & 970 & 75.30 & 930 \\ 
        200    & 92.78 & 990 & 72.47 & 880 \\ 
        1000   & 88.17 & 990 & 62.02 & 970 \\
        \bottomrule
    \end{tabular}
\end{table}


The experimental results for $\xi=100$ across varying client counts are demonstrated in Table~\ref{table:exp_by_clients}. The results show that as the total number of clients decreases, there is a corresponding increase in test accuracy. As the total number of clients decreases and the number of training samples is constant, the clients get larger training datasets during non-IID distribution. As the clients have larger local datasets with greater diversity, this results in better convergence of local models with increased accuracy.

\begin{table}[h]
\centering
    \caption{The best accuracies for the MNIST and CIFAR10 datasets across varying client numbers and privacy budgets.}
        \label{table:exp_by_eps}
    \begin{tabular}{llllll}
        \toprule
            & & \multicolumn{2}{c}{MNIST} & \multicolumn{2}{c}{CIFAR10} \\ 
            \midrule
            \# Clients & $\epsilon \times 10^3$ & Accuracy & Round & Accuracy & Round \\
            \midrule
            \multirow{3}{*}{50}     & 5   & 90.60 & 330 & 74.36 & 940   \\
                                    & 7.5 & 92.39 & 460 & 75.34 & 940   \\
                                    & 10  & 93.22 & 460 & 75.46 & 990   \\ 
            \midrule
            \multirow{3}{*}{100}    & 5   & 80.59 & 420 & 70.68 & 970   \\
                                    & 7.5 & 81.21 & 810 & 72.32 & 970   \\
                                    & 10  & 83.49 & 970 & 73.11 & 970   \\ 
            \midrule
            \multirow{3}{*}{200}    & 5   & 73.36 & 900 & 59.97 & 890   \\
                                    & 7.5 & 71.60 & 790 & 62.27 & 640   \\
                                    & 10  & 72.49 & 620 & 63.23 & 900   \\ 
            \midrule
            \multirow{6}{*}{1000}   & 5    & 63.97 & 820 & 22.00 & 10   \\
                                    & 7.5  & 65.39 & 990 & 24.72 & 20   \\
                                    & 10   & 66.50 & 760 & 31.19 & 40   \\
            \cmidrule{2-6}
                                    & 500  & \multirow{3}{*}{---} & \multirow{3}{*}{---} & 55.15 & 930   \\
                                    & 750  &  & & 58.02 & 930   \\
                                    & 1000 &  & & 59.35 & 920   \\ 
            \bottomrule
    \end{tabular}
\end{table}

We perform extensive experiments to find the right privacy budget $\epsilon$ while maximizing accuracy and preserving privacy. The results of experiments for varying privacy budgets and client counts are demonstrated in Table~\ref{table:exp_by_eps}. As the model complexity increases, the models get more sensitive to noise. Thus, we often require a high privacy budget, generating less noise. Moreover, the number of samples on clients ($n_i$) decreases as the client count increases, leading our sensitivity formulation in Eq.~\eqref{eq:dp_sens} to generate higher noise levels on the clients. This proves the increasing accuracy gap as the number of clients increases, considering baseline experiments presented in Table~\ref{table:exp_by_clients}. In the experiments with 1000 clients, we observe convergence failures for privacy budgets of $\epsilon \leq 10^3$ on the MNIST model and $\epsilon \leq 10^5$ on the CIFAR-10 model. For the CIFAR10, we find that privacy budgets of $\epsilon \geq 5 \times 10^5$ are more appropriate, as shown in Table~\ref{table:exp_by_eps}.

\begin{figure}[t]
    \centering
    \includegraphics[width=1\columnwidth]{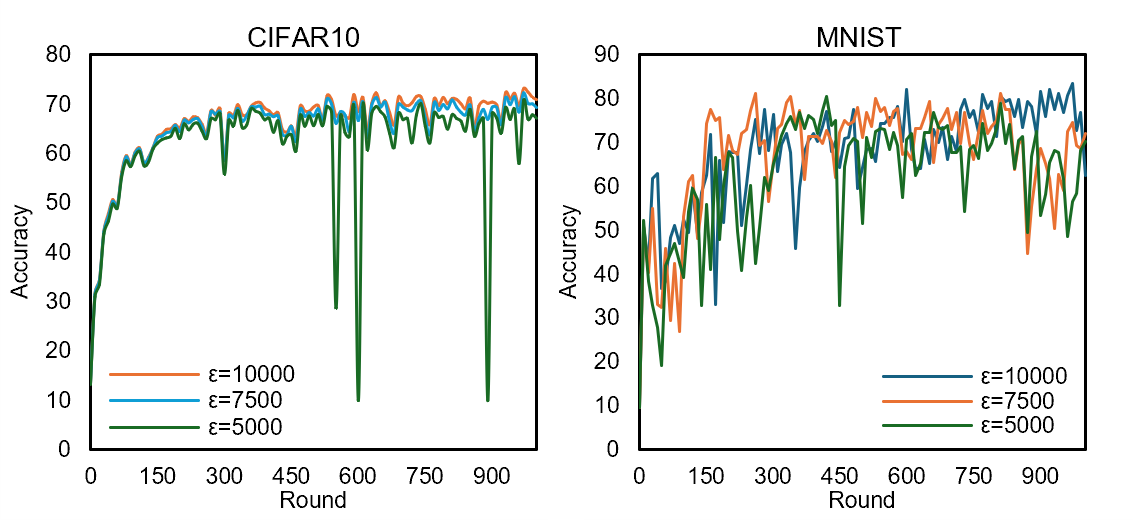}
    \caption{The test accuracy across different privacy budgets for 100 clients on the CIFAR10 and MNIST datasets. Best viewed in color.}
    \label{fig:exp_dp_training}
\end{figure}

The test accuracies during 1000 training rounds across different privacy budgets $\epsilon$ for 100 clients on the CIFAR10 and MNIST datasets are shown in Fig.~\ref{fig:exp_dp_training}. For the CIFAR10, higher privacy budgets ($\epsilon=10^4$) lead to better performance, with accuracy stabilizing around 70\%, while lower budgets ($\epsilon=5000$) result in a noticeable drop in accuracy. Similarly, for the MNIST, higher privacy budgets also yield better accuracy, with models reaching approximately 80\%. However, MNIST shows larger fluctuations, particularly in the early rounds, compared to CIFAR10, though it stabilizes with higher $\epsilon$ values. These results show the trade-off between privacy and accuracy.

\subsection{Adaptive Quantization} \label{sec_exp_adq}
In this section, we present a comprehensive evaluation of our adaptive quantization method, as described in Section~\ref{sec_quant}, across different client counts and bit-length schedulers. In our experiments, we apply the Laplacian-DP mechanism detailed in Section~\ref{sec_dp} with a privacy budget of $\epsilon=10^4$, except for the scenario with 1000 clients on CIFAR10, where $\epsilon=10^6$ is used. These privacy budgets are chosen because they offer a better balance between stability and accuracy compared to lower budget values, as described in Section~\ref{sec_exp_dp}. 

\begin{figure}[t]
    \centering
    \includegraphics[width=1\linewidth]{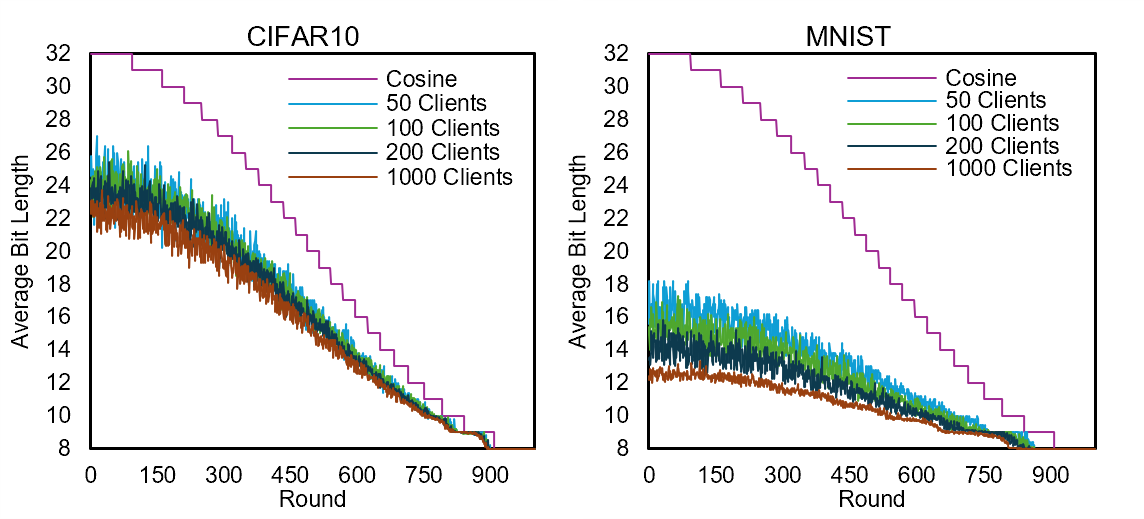}
    \caption{The average bit-length per round for varying client counts on CIFAR10 and MNIST datasets, where each client uses cosine annealing bit-length scheduling with client importance ($\lambda_h=0.75$, $b_{min}=8$). Best viewed in color.}
    \label{fig:exp_avg_bits}
\end{figure}

Our adaptive quantization approach includes a global bit-length scheduler using cosine annealing based on the number of communication rounds. During training, the server uses the global scheduler to quantize the global model before broadcasting. On the client side, we either apply the global bit-length directly or adjust it by weighting it according to the client's importance $\nu_i$, which is computed by the homogeneity and size of the local dataset. In our experiments, Cosine refers to using the global scheduler without incorporating client importance, while Dynamic refers to the global scheduler adjusted by the client importance. The average bit-lengths per round across varying numbers of clients, facilitated by these schedulers, are illustrated in Figure~\ref{fig:exp_avg_bits}.

\begin{figure*}[t]
    \centering
    \includegraphics[width=1\linewidth]{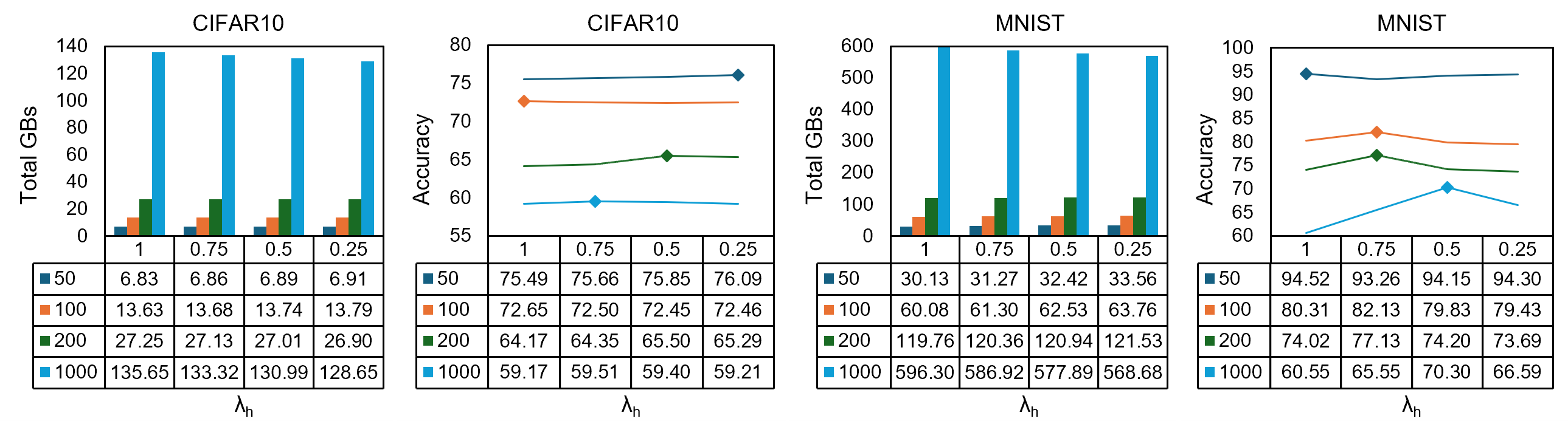}
    \caption{The best accuracies and total communicated gigabytes after 1000 training rounds for the CIFAR10 and MNIST datasets, evaluated across different $\lambda_h$ values and client counts, with a privacy budget of $\epsilon = 10^4$, except for the scenario with 1000 clients on CIFAR10, where $\epsilon = 10^6$ is used. Best viewed in color.}
    \label{fig:exp_lambdah}
\end{figure*}

We first analyze how varying $\lambda_h \in \{0.25, 0.5, 0.75, 1.0\}$ affects the accuracy and total communicated gigabytes across various client counts on non-IID datasets. The results of the experiments performed on CIFAR10 and MNIST datasets are shown in Figure~\ref{fig:exp_lambdah}. The highest accuracy achieved by corresponding $\lambda_h$ is marked as a diamond in the accuracy charts. As expected, increasing the number of clients per round leads to a significant rise in total communication costs across both datasets. For a large number of clients (e.g., 1000 clients) across both datasets, a decrease in $\lambda_h$ results in a slight reduction in total communicated gigabytes. In addition, the results indicate that the dataset size becomes more influential than homogeneity in terms of accuracy and total communicated gigabytes as the number of clients increases. For example, the lowest communication cost is achieved with $\lambda_h=0.25$ for both 200 and 1000 clients on CIFAR10, as well as for 1000 clients on MNIST. This is expected as the dataset size per client decreases with an increasing number of clients, making the dataset size more crucial than homogeneity.

\begin{table*}[t]
\centering
    \caption{The best accuracies and total communicated gigabytes after 1000 training rounds on the MNIST and CIFAR10 datasets, across varying client counts and bit-lengths with a privacy budget of $\epsilon = 10^4$, except for the scenario with 1000 clients on the CIFAR10, where $\epsilon = 10^6$ is used. \textit{Cosine} denotes cosine annealing without client importance ($b_{min} = 8$), while \textit{Dynamic} refers to cosine annealing with client importance ($\lambda_h = 0.75$, $b_{min} = 8$).}
    \label{table:exp_dpq_table}
    \begin{tabular}{llllllll}
        \toprule
        & & \multicolumn{3}{c}{MNIST} & \multicolumn{3}{c}{CIFAR10} \\ 
        \midrule
        Total Clients & Bit Length & Total GBs & Accuracy & Best Round & Total GBs & Accuracy & Best Round \\
        \midrule
        \multirow{6}{*}{50}     & INT4       & 7.75    & 76.65 & 860 & 1.52   & 50.56 & 340 \\ 
                                & INT8       & 15.49   & 92.85 & 460 & 3.03   & 75.76 & 900 \\
                                & INT16      & 30.98   & 94.61 & 610 & 6.07   & 75.74 & 900 \\
                                & FP32      & 61.97   & 93.22 & 460 & 12.13  & 75.46 & 990 \\ 
                                & Cosine  & 38.75   & 93.04 & 460 & 7.59   & 75.97 & 870 \\ 
                                & Dynamic & 31.27   & 93.26 & 460 & 6.86   & 75.66 & 900 \\ 
        \midrule 
        \multirow{6}{*}{100}    & INT4       & 15.49   & 60.49 & 380 & 3.03   & 46.31 & 140 \\ 
                                & INT8       & 30.98   & 81.50 & 730 & 6.07   & 72.70 & 970 \\ 
                                & INT16      & 61.97   & 78.08 & 340 & 12.13  & 72.67 & 970 \\
                                & FP32      & 123.93  & 83.49 & 970 & 24.27  & 73.11 & 970 \\
                                & Cosine  & 77.50   & 81.77 & 770 & 15.18  & 72.34 & 970 \\
                                & Dynamic & 61.30   & 82.13 & 890 & 13.68  & 72.50 & 640 \\
        \midrule 
        \multirow{6}{*}{200}    & INT4       & 30.98   & 49.37 & 230 & 6.07   & 41.74 & 50  \\ 
                                & INT8       & 61.97   & 76.49 & 540 & 12.13  & 65.24 & 790 \\ 
                                & INT16      & 123.93  & 74.96 & 710 & 24.27  & 64.85 & 510 \\
                                & FP32      & 247.86  & 72.49 & 620 & 48.53  & 63.23 & 900 \\
                                & Cosine  & 155.01  & 76.99 & 990 & 30.35  & 65.19 & 760 \\
                                & Dynamic & 120.36  & 77.13 & 540 & 27.13  & 64.35 & 890 \\
        \midrule
        \multirow{6}{*}{1000}   & INT4       & 154.91  & 40.17 & 450 & 30.33  & 34.41 & 240 \\ 
                                & INT8       & 309.83  & 64.40 & 990 & 60.67  & 59.33 & 999 \\ 
                                & INT16      & 619.65  & 65.76 & 980 & 121.34 & 59.49 & 999 \\
                                & FP32      & 1239.31 & 66.50 & 760 & 242.67 & 59.35 & 920 \\
                                & Cosine  & 775.03  & 64.85 & 820 & 151.76 & 59.54 & 920 \\
                                & Dynamic & 586.92  & 65.55 & 610 & 133.32 & 59.51 & 930 \\
        \bottomrule
    \end{tabular}
\end{table*}

Moreover, the highest accuracy is achieved with $\lambda_h = 0.5$ for 1000 clients on MNIST. The results on CIFAR10 indicate that $\lambda_h$ has a minimal effect on accuracy across varying numbers of clients, leading to a maximum change of only 1.12\%, observed for 200 clients. In contrast, for MNIST, we observe a significantly larger impact, with a maximum change of 9.75\%, observed for 1000 clients. In summary, for both datasets, across varying numbers of clients, $\lambda_h = 0.75$ or $\lambda_h = 0.5$ balances accuracy and communication cost. For example, the highest accuracy values are achieved with $\lambda_h = 0.75$ for 100 and 200 clients on MNIST and for 1000 clients on CIFAR10. Therefore, we use $\lambda_h = 0.75$ in our subsequent experiments.

\begin{figure*}[t]
    \centering
    \includegraphics[width=1\linewidth]{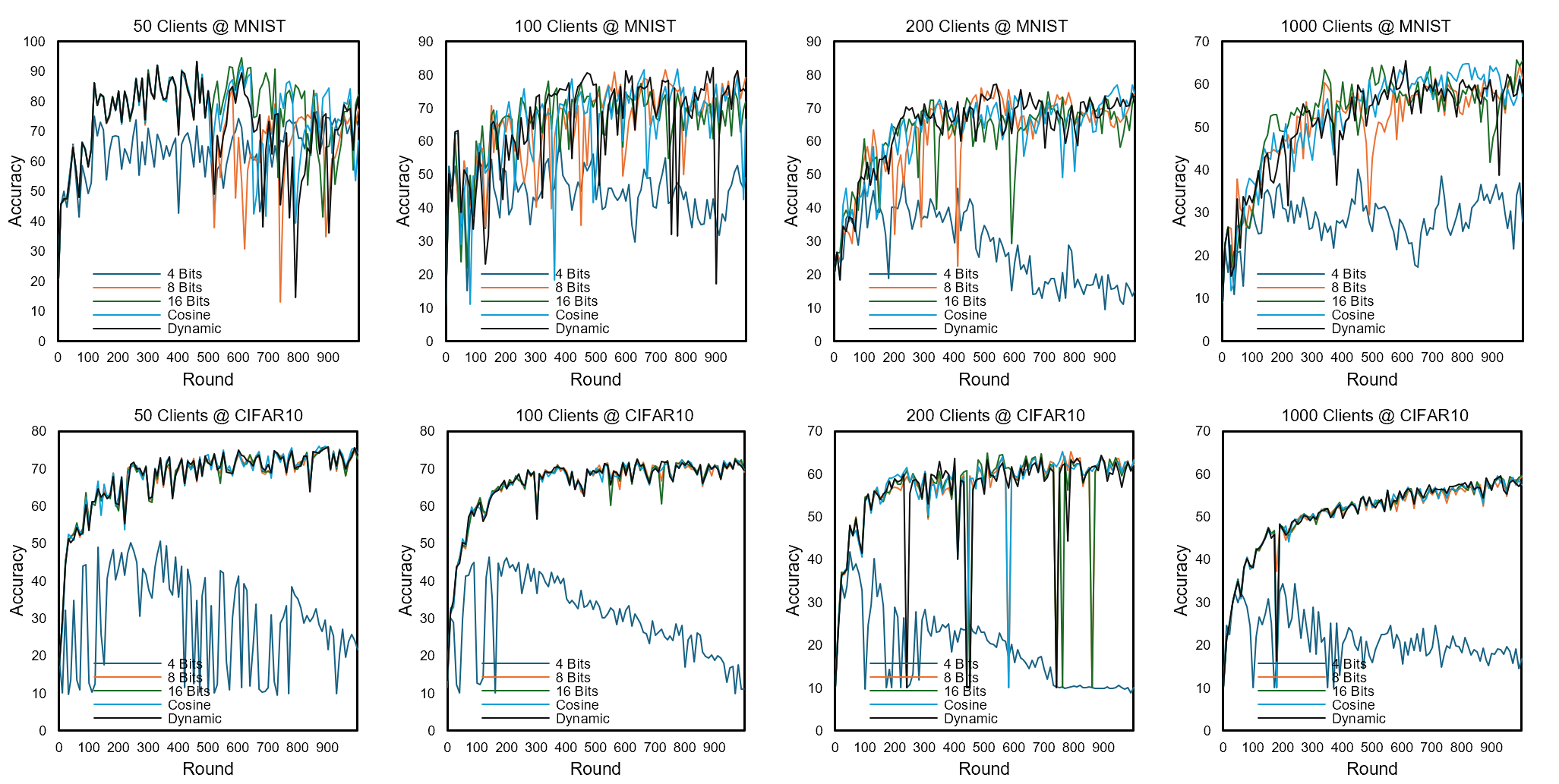}
    \caption{The test accuracies for the MNIST and CIFAR10 datasets across varying client counts and bit-lengths, with a privacy budget of $\epsilon = 10^4$, except for the scenario with 1000 clients on CIFAR10, where $\epsilon = 10^6$ is used. \textit{Cosine} denotes cosine annealing without client importance ($b_{min} = 8$), while \textit{Dynamic} refers to cosine annealing with client importance ($\lambda_h = 0.75$, $b_{min} = 8$). Best viewed in color.}
    \label{fig:exp_dpq_acc}
\end{figure*}

We perform a comprehensive evaluation of our bit-length scheduling methods, Cosine and Dynamic, across various client counts and non-IID datasets. We also use static quantization with 4, 8 and 16 bits to better analyze the trade-off between accuracy and communication. We include the results of full precision (32-bit float) training to better illustrate the impact of quantization on both model performance and communication cost. The experimental results, including the best accuracy observed over 1000 rounds and the total communicated gigabytes accumulated across all rounds, are presented in Table~\ref{table:exp_dpq_table}. Furthermore, the global test accuracies, computed every 10 rounds on the server for varying numbers of clients and both datasets, are illustrated in Figure~\ref{fig:exp_dpq_acc}. 

The results show that lower bit-lengths reduce communication costs but come at the expense of accuracy, especially in the 4-bit settings where severe quantization errors obscure model updates and introduce training instability, particularly with non-IID data, as illustrated in Figure~\ref{fig:exp_dpq_acc}. Across all numbers of clients, the 8-bit quantization emerges as a practical choice, providing a favorable balance between accuracy and communication cost. The Cosine and Dynamic methods further optimize this balance by maintaining similar or slightly better accuracies while reducing the total communicated gigabytes. The Cosine method provides about 37.46\% reduction in total communicated gigabytes compared to the 32-bit setting. The Dynamic method achieves even greater efficiency, nearly halving communication cost, with reductions ranging from 49.54\% to 52.64\% on MNIST and from 43.45\% to 45.06\% on CIFAR10 across 50 to 1000 clients.

For 50 clients, the Cosine method achieves an accuracy of 93.04\% on MNIST and 75.97\% on CIFAR10, with total communication costs of 38.75 GBs and 7.59 GBs, respectively. This is more efficient than the full precision setting, which delivers similar accuracy but incurs significantly higher communication costs of 61.97 GB for MNIST and 12.13 GB for CIFAR10. The Dynamic method further improves efficiency by reducing the communication to 31.27 GBs for MNIST and 6.86 GBs for CIFAR10, maintaining comparable accuracies of 93.26\% and 75.66\%, respectively. For MNIST, the 16-bit static quantization achieves the highest accuracy of 94.61\% while reducing communication costs by 50\% compared to the full precision, likely due to the regularization effect introduced by the information loss during quantization.

For 100 clients, the effectiveness of our methods becomes more evident. The Dynamic method achieves the highest accuracy of 82.13\% on MNIST with a communication cost of 61.30 GBs, exceeding the 16-bit setting that achieves 78.08\% accuracy with slightly higher communication. On CIFAR10, both Cosine and Dynamic methods maintain competitive accuracies around 72\%, with reduced communication cost by nearly 50\% compared to the 32-bit setting. The 8-bit quantization maintains competitive accuracies of 72.70\% on CIFAR10 and 81.50\% on MNIST compared to the 32-bit setting while reducing the communication cost by a factor of four, showing surprisingly efficient performance at these client counts.

For 200 clients, the Cosine and Dynamic methods continue to deliver better trade-offs between accuracy and communication cost. The Dynamic method achieves the highest accuracy of 77.13\% on MNIST with a total of 120.36 GBs communicated, which is more efficient than the 32-bit setting that communicates 247.86 GBs for a lower accuracy of 72.49\%. For CIFAR10, the Cosine method achieves the highest accuracy of 65.19\%, while the Dynamic method has 64.35\% accuracy, which is still higher than the 32-bit training. However, the Dynamic method with $\lambda_h = 0.5$ achieves an accuracy of 65.50\%, which is the highest one at these client counts on CIFAR10. This trend shows the increasing influence of dataset size as the number of clients increases. Besides, the 8-bit quantization achieves competitive accuracies around 76\% for MNIST and 65\% for CIFAR10 while reducing communication costs by a factor of four in both cases.

With 1000 clients, where communication efficiency becomes essential at such a large scale, our methods continue to demonstrate their scalability and effectiveness. The Dynamic method achieves an accuracy of 65.55\% on MNIST and 59.51\% on CIFAR10 while reducing communication costs by 52.64\% for MNIST and 45.06\% for CIFAR10, all while maintaining accuracy levels comparable to the full precision training. For CIFAR10, the Cosine scheduler achieves the highest accuracy of 59.54\%, but with a slightly higher communication cost compared to the Dynamic method. For MNIST, however, the Dynamic scheduler performs better than the Cosine scheduler in terms of accuracy and communication cost, proving the effectiveness of the client importance strategy at these client counts. Additionally, the 8-bit and 16-bit static quantization schemes demonstrate their effectiveness at these client counts for both datasets, offering a balanced approach that maintains competitive accuracy while significantly reducing communication costs.

In summary, our adaptive quantization methods, including the Cosine and Dynamic schedulers, successfully balance accuracy and communication efficiency, demonstrating their scalability in large-scale FL. By leveraging a global bit-length scheduler based on cosine annealing and introducing entropy-based client importance, our methods reduce communication costs by over 50\% on MNIST and 45\% on CIFAR10 compared to 32-bit setting while maintaining competitive accuracy. Even with 1000 clients, where communication efficiency is crucial, our Dynamic scheduler delivers superior performance, significantly reducing communication overhead with minimal accuracy loss. Although static quantization methods, especially 8-bit, are shown to be effective, they require selecting a fixed bit precision value in advance, which can make them difficult to apply in real-world problems. In contrast, our adaptive quantization methods offer greater flexibility by dynamically adjusting the precision based on the number of communication rounds or the characteristics of local datasets, making them more adaptable and practical for varying conditions. 

One of the key advantages of our methods is their ease of implementation, making them accessible for a wide range of applications. Additionally, the entropy calculation used in our approach has linear time complexity, ensuring that the methods are both fast and efficient. Furthermore, they are highly adaptable, allowing them to be applied to any problem without the need for extensive modifications, which enhances their practicality in diverse scenarios. However, our Dynamic scheduler includes the $\lambda_h$ parameter, which may require fine-tuning for optimal performance. Additionally, our methods do not take into account the quality of the samples on the clients, which could be considered a potential limitation. This can be alleviated by using client or data valuation methods, which estimate the contribution of client or data to the model performance \cite{wang2020principled, li2021sample, eardic_fl, ardic_ss_fl}.

Importantly, although the clients in our algorithm introduce noise in two stages, first by Laplacian-based DP and then by stochastic uniform quantization, this two-step process preserves the unbiased nature of the uploaded information. Specifically, our DP mechanism adds noise drawn from a Laplace distribution centered at zero. The zero-mean property ensures that, on average, the expected value of the noise is zero, so it does not shift the true value of the uploaded information. Similarly, our adaptive quantization methods employ stochastic uniform quantization, in which each real value is probabilistically rounded either up or down so that the expected quantized value equals the original value, as also proven by Alistarh et al. \cite{alistarh2017qsgd}. Consequently, since both the noise additions are unbiased, one by its zero-mean and the other by its probabilistic rounding, the combination of these two methods does not introduce systematic bias into the model updates, even though it results in a higher variance.

\subsection{Performance on Medical Image Datasets} \label{sec_exp_health}

To further evaluate the effectiveness of our adaptive quantization methods, we perform experiments on widely used medical image classification datasets: PAP-Smear \cite{papsmear}, Chest X-ray (Pneumonia) \cite{octr_chxray}, and BreakHisV1 \cite{breakhisv1}. The details of these datasets are shown in Table~\ref{table:mid_datasets}. We aim to provide a comprehensive analysis of the trade-offs between communication efficiency and model performance in FL. We extensively evaluate our methods using metrics such as total communicated gigabytes, accuracy, precision, recall, F1 score, and balanced accuracy score (BACC).

\begin{table}[h]
    \centering
	\caption{The details of the medical image classification datasets.}
        \label{table:mid_datasets}
	\begin{tabular}{llll}
		\toprule
            Dataset & PAP-Smear & Pneumonia & BreakHisV1 \\
		\midrule
            Image Size          & 128x128     & 224x224     & 128x128 \\
            Train Samples       & 3,049       & 5,232       & 5,361   \\
            Test Samples        & 1,000       & 624         & 2,548   \\
            \# Classes          & 5           & 2           & 2       \\
		\bottomrule
	\end{tabular}
\end{table}


We utilize the ImageNet pre-trained EfficientNet-B0 model within FedML \cite{fedml2020}, which is well-suited for medical image classification due to its efficient parameterization and strong feature extraction capabilities. We use the FedAvg algorithm with 10 clients over 100 communication rounds and adopt the Adam optimizer with a learning rate of $3\times10^{-4}$ and weight decay of $5\times10^{-4}$. The batch size is 64, and each client performs two epochs of local training. Every three rounds, we measure the performance metrics on the central server. Throughout the training process, we monitor the number of rounds that yield the highest test BACC, computed on the server using the global model, referred to as the "Best Round" in experiments. Additionally, we maintain a fixed privacy budget of \(\epsilon = 10^4\) and gradient norm of $\xi=1000$ for the Laplacian-based DP, as our experiments have demonstrated that these values provide an optimal balance between privacy preservation and model performance.

For the non-IID distribution of the datasets, we use a Dirichlet distribution with an alpha parameter of 0.5, as in the CIFAR10 and MNIST experiments. We evaluate our bit-length scheduling methods, Cosine and Dynamic, on these non-IID datasets. To better analyze the trade-off between accuracy and communication, we employ static quantization with 12 and 16 bits. A minimum bit length of \(b_{\text{min}} = 12\) is chosen as the model collapses with lower values. For the Dynamic method, we set \(\lambda_h = 0.75\) and \(b_{\text{min}} = 12\). We include the results of full precision (32-bit float) training to illustrate better the impact of quantization on both model performance and communication cost. The experimental results, including the best BACC, observed over 100 rounds and the total communicated gigabytes accumulated across all rounds, are presented in Table~\ref{table:exp_by_mid}. 

\begin{table*}[t]
\centering
    \caption{The best performances and total communicated gigabytes after 100 training rounds on the medical image datasets for 10 clients across varying bit-lengths with a privacy budget of $\epsilon = 10^4$. \textit{Cosine} denotes cosine annealing without client importance ($b_{min} = 12$), while \textit{Dynamic} refers to cosine annealing with client importance ($\lambda_h = 0.75$, $b_{min} = 12$).}
        \label{table:exp_by_mid}
    \begin{tabular}{llllllllll}
        \toprule
            Dataset & Bit Length & Total GBs & Accuracy & Precision & Recall & F1 Score & BACC & Best Round \\
            \midrule
            \multirow{5}{*}{PAP-Smear}  & INT12      & 11.33 & 86.20 & 87.88 & 86.28 & 86.31 & 86.28 & 75 \\
                                        & INT16      & 15.11 & 90.90 & 91.12 & 91.03 & 90.94 & 91.03 & 81 \\
                                        & FP32      & 30.22 & 89.90 & 90.72 & 90.07 & 89.96 & 90.07 & 87 \\
                                        & Cosine  & 20.85 & 88.80 & 90.18 & 88.99 & 88.95 & 88.99 & 87 \\
                                        & Dynamic & 19.00 & 89.40 & 90.21 & 89.59 & 89.41 & 89.59 & 81 \\
            \midrule
            \multirow{5}{*}{Pneumonia}  & INT12      & 11.33 & 92.95 & 93.40 & 91.54 & 92.32 & 91.54 & 30 \\
                                        & INT16      & 15.11 & 91.51 & 93.48 & 88.93 & 90.50 & 88.93 & 30 \\
                                        & FP32      & 30.22 & 94.39 & 94.19 & 93.80 & 93.99 & 93.80 & 33 \\
                                        & Cosine  & 20.85 & 94.23 & 94.22 & 93.42 & 93.79 & 93.42 & 27 \\
                                        & Dynamic & 18.73 & 92.63 & 93.27 & 91.03 & 91.94 & 91.03 & 27 \\
            \midrule
            \multirow{5}{*}{BreakHisV1} & INT12      & 11.33 & 81.00 & 80.42 & 82.27 & 80.57 & 82.27 & 72 \\
                                        & INT16      & 15.11 & 90.23 & 89.28 & 90.64 & 89.80 & 90.64 & 84 \\
                                        & FP32      & 30.22 & 91.95 & 91.43 & 91.45 & 91.44 & 91.45 & 99 \\
                                        & Cosine  & 20.85 & 86.70 & 86.18 & 88.39 & 86.40 & 88.39 & 72 \\
                                        & Dynamic & 18.43 & 87.44 & 86.52 & 88.36 & 87.03 & 88.36 & 72 \\
            \bottomrule
    \end{tabular}
\end{table*}

\begin{figure*}[t]
    \centering
    \includegraphics[width=1\linewidth]{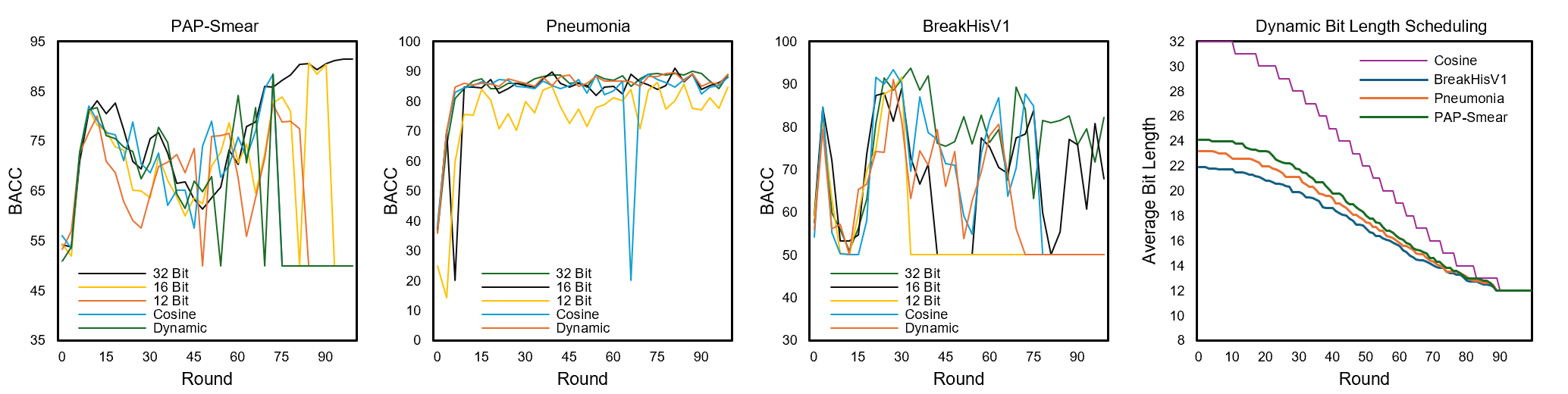}
    \caption{The BACC scores on the global test dataset and average bit-lengths per round for the medical imaging datasets under different bit-length settings, with 10 clients and a privacy budget of $\epsilon = 10^4$. The average bit-length is computed as the mean bit-length across all clients. \textit{Cosine} denotes cosine annealing without client importance ($b_{min} = 12$), while \textit{Dynamic} refers to cosine annealing with client importance ($\lambda_h = 0.75$, $b_{min} = 12$). Best viewed in color.}
    \label{fig:exp_dpq_mid_bacc}
\end{figure*}

The results reveal that both Cosine and Dynamic methods significantly reduce the total communication cost compared to the 32-bit baseline while maintaining competitive performance in terms of F1 Score and BACC. On the PAP-Smear dataset, the Dynamic method achieves a 37\% reduction in communication with only a minimal decrease in F1 Score from 90.94\% to 89.41\% and BACC from 91.03\% to 89.59\%. For the Pneumonia dataset, the Cosine method preserves near-baseline performance with an F1 Score of 93.79\% compared to 93.99\% for the 32-bit baseline while reducing communication by 31\%. Similarly, on the BreakHisV1 dataset, the Dynamic method reduces the communication cost by approximately 39\% compared to the 32-bit baseline, with only a modest reduction in F1 Score from 89.80\% to 87.03\% and BACC from 90.64\% to 88.36\%.

In some cases, the 16-bit static integer quantization not only results in higher model performance but also has lower total communicated gigabytes compared to our Cosine and Dynamic methods. For instance, on the PAP-Smear dataset, the 16-bit quantization achieves higher F1 and BACC scores than both the Cosine and Dynamic methods while also communicating less data. This suggests that a fixed moderate bit-length can sometimes offer a better trade-off between communication efficiency and model performance. This outcome can be attributed to the adaptive nature of our Cosine and Dynamic methods, which, while designed to optimize communication by dynamically adjusting bit-lengths, may occasionally use higher bit-lengths during certain training rounds, leading to increased communication overhead compared to a fixed 16-bit setting. Additionally, the dynamic bit-length adjustment may introduce greater variability in model performance due to fluctuations in quantization noise. This can be observed in Figure~\ref{fig:exp_dpq_mid_bacc}, where the BACC scores on the test dataset show greater fluctuation compared to the 32-bit float setting.

These findings highlight the nuanced trade-off between communication efficiency and model performance inherent in quantization methods. While the 16-bit static quantization offers better performance and lower communication cost in some cases due to its fixed, moderate precision, our adaptive methods provide flexibility by adjusting bit-lengths based on training dynamics and client importance. This flexibility becomes increasingly valuable as the number and diversity of clients grow, making our Cosine and Dynamic methods particularly effective for large-scale FL environments with fluctuating client participation. However, this adaptability may not always result in superior performance or lower communication overhead compared to a well-chosen fixed bit-length configuration. Moreover, the combination of Laplacian noise and quantization error poses significant challenges to maintaining model accuracy and stability, as shown in Figure~\ref{fig:exp_dpq_mid_bacc}. Laplacian noise, introduced for privacy preservation, adds randomness to the training process, while quantization error reduces numerical precision. When these factors interact, they can amplify performance variability, complicating the balance between efficient communication and reliable model performance. Therefore, although adaptive quantization methods offer promising flexibility, their success depends heavily on careful management of noise and precision trade-offs.

\section{Conclusion}

In this paper, we proposed a novel FL approach that combines adaptive quantization with Laplacian-based DP to improve both communication efficiency and privacy across varying numbers of clients and non-IID datasets. We employed Laplacian-based DP to preserve privacy, which is relatively underexplored in FL across varying numbers of clients and offers tighter privacy guarantees than Gaussian-based DP. Additionally, we proposed a simple and efficient global bit-length scheduler using round-based cosine annealing, along with a client-based scheduler that dynamically adapts based on client contribution. We evaluated our method through extensive experiments on CIFAR10 and MNIST datasets across various client counts, bit-length schedulers, and privacy budgets. The experimental results demonstrated that our adaptive quantization methods reduced communication costs by up to 52.64\% on MNIST and 45.06\% on CIFAR10 compared to the 32-bit float setting while preserving model accuracy even in the presence of privacy noise.

To further validate the effectiveness of our approach, we conducted experiments on medical image datasets, including PAP-Smear, Pneumonia (Chest X-ray), and BreakHisV1, which present unique challenges due to their complex image structures and critical classification requirements. Our adaptive quantization methods demonstrated a significant reduction in communication overhead, ranging from 31\% to 37\%, while maintaining competitive accuracy levels compared to 32-bit float settings. Notably, the Cosine and Dynamic bit-length schedulers achieved a strong balance between efficiency and performance, proving their applicability in real-world healthcare scenarios where communication and privacy constraints are crucial.

Future work will focus on refining the adaptive quantization methods by exploring more advanced algorithms for accurately estimating client importance, aiming to further enhance the balance between communication efficiency and model performance in large-scale FL. We also plan to investigate the integration of more advanced privacy mechanisms, such as secure multiparty computation, to further strengthen privacy guarantees in FL. These enhancements aim to further optimize communication and privacy trade-offs, making FL more applicable to real-world problems.

\section*{Acknowledgment}

We sincerely thank Prof. Fatih Erdoğan Sevilgen and Assoc. Prof. Mehmet Göktürk for their valuable insights and constructive feedback. Their guidance has significantly contributed to the quality of this work.

\bibliographystyle{IEEEtran}
\balance
\bibliography{fl-refs}

\begin{IEEEbiography}[{\includegraphics[width=1in,height=1.25in,clip,keepaspectratio]{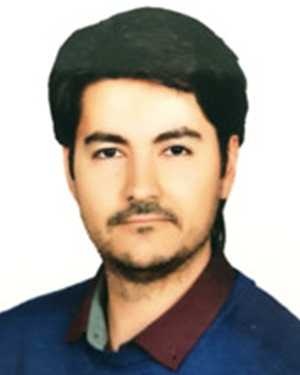}}]{Emre ARDIÇ} received the B.S. and M.S. degrees in computer engineering from
Gebze Technical University, Kocaeli, Turkey, in 2014 and 2018, respectively. Since 2018, he has been pursuing his Ph.D. in computer engineering at Gebze Technical University, where his research centers on federated learning, communication efficiency, and privacy-preserving techniques in distributed systems. Since 2014, he has been a Researcher at The Scientific and Technological Research Council of Turkey (TÜBİTAK), focusing on various projects related to deep learning. His interests also include image processing, adaptive quantization, and differential privacy in non-IID data environments.

\end{IEEEbiography}
\vskip -2\baselineskip plus -1fil
\begin{IEEEbiography}[{\includegraphics[width=1in,height=1.25in,clip,keepaspectratio]{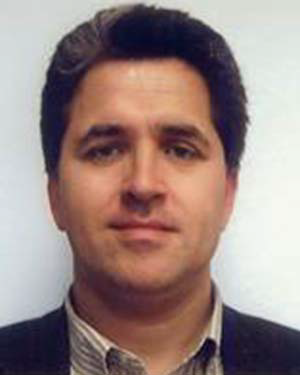}}]{Yakup GENÇ} received the Ph.D. degree in computer science from the University of Illinois at Urbana-Champaign, Champaign, IL, USA. In September 1999, he joined Siemens Corporate Research (SCR). As a Research Scientist, Project Manager, Program Manager, and Group Manager, he developed technology and research strategies in the areas of computer vision, augmented reality, and machine learning. His tenure with SCR produced numerous publications and patents. Since September 2012, he has been a Member of the Faculty of the Computer Engineering Department at Gebze Technical University, Gebze, Türkiye, where he continuous to conduct research in fields of computer vision, augmented reality, autonomous vehicles, machine/deep learning while maintaining close ties with industry for practical applications of his research.
\end{IEEEbiography}

\EOD

\end{document}